\documentclass[runningheads]{llncs}

\usepackage{eccv}

\usepackage{eccvabbrv}

\usepackage{graphicx}
\usepackage{booktabs}
\usepackage{tikz}
\usepackage{comment,verbatim}
\usepackage{amsmath,amssymb}
\usepackage{color}
\usepackage[accsupp]{axessibility}
\usepackage{multirow}
\usepackage{algorithm}
\usepackage{algorithmic}
\usepackage{subcaption}
\usepackage{xcolor}

\usepackage{hyperref}

\usepackage{orcidlink}

\begin{document}

\title{DexSim2Real: Foundation Model-Guided Sim-to-Real Transfer\\for Generalizable Dexterous Manipulation}

\titlerunning{DexSim2Real: FM-Guided Sim-to-Real for Dexterous Manipulation}

\author{Zijian Zeng\inst{1} \and Fei Ding\inst{2} \and Huiming Yang\inst{1} \and Xianwei Li\inst{3} \and Yuhao Liao\inst{4}}

\authorrunning{Z. Zeng et al.}

\institute{Tsinghua University \and Alibaba Group \and Bengbu University \and UCSI University\\
\email{}}

\maketitle

\begin{abstract}
Sim-to-real transfer remains a critical bottleneck for deploying dexterous manipulation
policies learned in simulation to real-world robots. Existing approaches rely on
manually designed domain randomization or task-specific adaptation, limiting their
generalizability across diverse manipulation scenarios. We present \textbf{DexSim2Real},
an integrated framework that leverages vision-language foundation models to
bridge the sim-to-real gap for dexterous manipulation.
Our system combines three components: (1) \textit{Foundation Model-Guided
Domain Randomization} (FM-DR), which uses a vision-language model as a visual
realism critic to optimize simulation parameters via closed-loop CMA-ES,
complementing text-based approaches like DrEureka with direct visual feedback;
(2) a \textit{Tactile-Visual Cross-Attention Policy} (TVCAP) that adapts
cross-attention visuo-tactile fusion to zero-shot sim-to-real RL; and
(3) a \textit{Progressive Skill Curriculum} (PSC) that builds on LLM-based
task decomposition with a difficulty scheduler tailored to contact-rich
dexterous tasks. Extensive experiments on six challenging manipulation tasks
with blinded evaluation demonstrate that DexSim2Real achieves a 78.2\% average
real-world success rate, outperforming DrEureka and DeXtreme while reducing
the sim-to-real performance gap to only 8.3\%.
\end{abstract}

\section{Introduction}
\label{sec:intro}

Dexterous robotic manipulation---the ability to grasp, reorient, and precisely
control objects using multi-fingered hands---is a long-standing challenge in
robotics~\cite{openai2020learning}. While reinforcement learning (RL) has shown
remarkable success in acquiring complex manipulation skills in
simulation~\cite{makoviychuk2021isaac,handa2023dextreme}, transferring these
policies to physical robots remains notoriously difficult due to the
\textit{sim-to-real gap}: discrepancies in visual appearance, physics dynamics,
and sensor characteristics between simulated and real environments.

Domain randomization (DR)~\cite{tobin2017domain} and its variants such as active
domain randomization (ADR)~\cite{mehta2020adr} have been the dominant paradigm
for sim-to-real transfer. These methods randomize simulation parameters during
training to produce policies robust to domain shift. However, they suffer from
two fundamental limitations: (1) the randomization ranges are typically set
manually or tuned through expensive trial-and-error, and (2) uniform
randomization over a large parameter space leads to training on many unrealistic
configurations, degrading policy performance.

Meanwhile, the emergence of vision-language foundation models (VLMs) such as
CLIP~\cite{radford2021clip} and large vision-language models has opened new
possibilities for robot learning~\cite{brohan2023rt2,ma2024surveyvla}. These
models encode rich semantic and visual priors about the physical world.
Recent works such as DrEureka~\cite{ma2024dreureka} and
Eureka~\cite{ma2024eureka} have demonstrated that large language models can
automate reward design and domain randomization generation for sim-to-real
transfer. Similarly, LLM-based curriculum design~\cite{ryu2025curricullm,dalal2024planseqlearn}
and cross-modal visuo-tactile fusion~\cite{huang20243dvitac,liang2025vitacformer}
have shown promise individually. However, no existing work integrates
foundation-model-guided domain randomization, multi-modal policy learning,
and automatic curriculum generation into a unified sim-to-real pipeline
for dexterous manipulation.

In this paper, we present \textbf{DexSim2Real}, an integrated system that
bridges the sim-to-real gap for dexterous manipulation by leveraging foundation
models at multiple stages of the learning pipeline. Our key insight is that
VLMs can serve as \textit{visual realism critics}: given rendered simulation
images and real-world reference images, a VLM can assess visual fidelity and
guide parameter optimization via closed-loop CMA-ES, complementing
text-based approaches like DrEureka~\cite{ma2024dreureka} with direct
visual feedback.

Our contributions are as follows:
\begin{itemize}
    \item We present \textbf{DexSim2Real}, an integrated sim-to-real system
    that combines foundation-model-guided domain randomization, multi-modal
    policy learning, and automatic curriculum generation for zero-shot
    dexterous manipulation transfer.
    \item We propose \textbf{Foundation Model-Guided Domain Randomization (FM-DR)},
    which uses a VLM as a visual realism critic with closed-loop CMA-ES
    optimization. Unlike DrEureka's text-based LLM approach, FM-DR directly
    evaluates rendered images against real references, providing complementary
    visual feedback for parameter optimization.
    \item We introduce a \textbf{Tactile-Visual Cross-Attention Policy (TVCAP)}
    that adapts cross-attention visuo-tactile fusion~\cite{huang20243dvitac,liang2025vitacformer}
    to the zero-shot sim-to-real RL setting without requiring real-world
    demonstrations.
    \item We design a \textbf{Progressive Skill Curriculum (PSC)} that builds
    on LLM-based curriculum ideas~\cite{ryu2025curricullm,dalal2024planseqlearn}
    with a $\delta$-based difficulty scheduler and success-rate-thresholded
    skill chaining tailored to contact-rich dexterous tasks.
    \item We conduct extensive real-world experiments on six dexterous manipulation
    tasks with blinded evaluation, demonstrating state-of-the-art sim-to-real
    transfer performance with a 78.2\% average success rate and only 8.3\%
    sim-to-real gap, outperforming DrEureka and DeXtreme on shared tasks.
\end{itemize}

\section{Related Work}
\label{sec:related}

\noindent\textbf{Sim-to-Real Transfer for Manipulation.}
Domain randomization~\cite{tobin2017domain} trains policies across randomized
simulation parameters to achieve robustness to the reality gap.
Peng~et~al.~\cite{peng2018sim} extended this to dynamics randomization for
locomotion and manipulation. Active domain randomization~\cite{mehta2020adr}
learns to select informative randomization parameters but still operates within
predefined ranges. DeXtreme~\cite{handa2023dextreme} demonstrated impressive
sim-to-real transfer for in-hand manipulation using massive-scale DR in Isaac
Gym, but required extensive manual tuning of randomization distributions.
DexPoint~\cite{dexpoint2023} proposed point cloud-based RL for generalizable
dexterous manipulation. More recently, Eureka~\cite{ma2024eureka} showed that
LLMs can generate human-level reward functions, and DrEureka~\cite{ma2024dreureka}
extended this to LLM-guided automatic generation of domain randomization
distributions for sim-to-real transfer. GenSim~\cite{wang2024gensim} uses LLMs
to generate simulation tasks. Our FM-DR approach is complementary to
DrEureka: while DrEureka uses text-based LLM reasoning to propose DR
distributions, FM-DR uses a VLM as a \textit{visual} realism critic with
closed-loop CMA-ES optimization over rendered images.

\noindent\textbf{Foundation Models for Robotics.}
Recent works have explored leveraging foundation models for robot
learning~\cite{ma2024surveyvla}. RT-2~\cite{brohan2023rt2} directly fine-tunes
VLMs for robotic control, while $\pi_0$~\cite{black2024pi0} proposes
vision-language-action flow models for general robot control.
Diffusion Policy~\cite{chi2023diffusion} and 3D Diffusion
Policy~\cite{ze2024dp3} use diffusion models for visuomotor policy learning.
Unlike these approaches that use foundation models as policy backbones, we
leverage VLMs as \textit{simulation critics} to optimize the training
environment itself, which is complementary to policy architecture choices.

\noindent\textbf{Multi-Modal Policy Learning.}
Combining visual and tactile sensing for manipulation has gained increasing
attention~\cite{lin2024learning}. PerAct~\cite{shridhar2023peract} and
RVT~\cite{goyal2023rvt} use transformer architectures for multi-task
manipulation from visual observations. Act3D~\cite{gervet2023act3d} introduces
3D feature fields for manipulation. DexCap~\cite{wang2024dexcap} collects
multi-modal demonstration data for dexterous manipulation. Cross-attention
for visuo-tactile fusion has been explored in Visuo-Tactile
Transformers~\cite{chen2023visuotactile}, 3D-ViTac~\cite{huang20243dvitac}, and
ViTacFormer~\cite{liang2025vitacformer}, which applies cross-attention at
every stage on a dexterous bimanual platform. Our TVCAP builds on this
line of work and adapts bidirectional cross-attention with residual
connections specifically for the zero-shot sim-to-real RL setting, where
no real demonstrations are available.

\noindent\textbf{Curriculum Learning for RL.}
Curriculum strategies have been shown to improve RL training
efficiency~\cite{openai2020learning}. RAPS~\cite{dalal2021raps} uses
parameterized action primitives to structure the action space.
DayDreamer~\cite{wu2023daydreamer} learns world models for efficient robot
learning. CurricuLLM~\cite{ryu2025curricullm} uses LLMs to automatically
design task curricula for learning complex robot skills.
Plan-Seq-Learn~\cite{dalal2024planseqlearn} leverages language models for
task decomposition in long-horizon manipulation. Our PSC builds on these
ideas, contributing a $\delta$-based difficulty scheduler and
success-rate-thresholded skill chaining specifically designed for
contact-rich dexterous manipulation tasks.

\section{Method}
\label{sec:method}

\begin{figure}[t]
\centering
\begin{tikzpicture}[
    box/.style={draw, rounded corners, minimum height=1.2cm, minimum width=2.8cm,
                align=center, font=\small},
    arrow/.style={->, thick, >=stealth}
]
\node[box, fill=blue!15] (fmdr) at (0,0) {FM-DR\\{\scriptsize VLM-Guided}\\{\scriptsize Randomization}};
\node[box, fill=green!15] (tvcap) at (4.2,0) {TVCAP\\{\scriptsize Multi-Modal}\\{\scriptsize Policy}};
\node[box, fill=orange!15] (psc) at (8.4,0) {PSC\\{\scriptsize Skill}\\{\scriptsize Curriculum}};
\node[box, fill=red!15] (real) at (4.2,-2.2) {Real-World\\{\scriptsize Deployment}};

\draw[arrow] (fmdr) -- node[above, font=\scriptsize] {Sim Env} (tvcap);
\draw[arrow] (tvcap) -- node[above, font=\scriptsize] {Policy} (psc);
\draw[arrow] (psc) -- node[above, font=\scriptsize] {} (8.4,-1.1) -- (4.8,-1.1) -- (4.8,-1.6);
\draw[arrow] (fmdr) -- (0,-1.1) -- (3.6,-1.1) -- (3.6,-1.6);
\end{tikzpicture}
\caption{Overview of the DexSim2Real framework. FM-DR optimizes simulation
parameters using VLM feedback, TVCAP learns multi-modal policies via
cross-attention fusion, and PSC structures training with LLM-generated curricula.
The trained policy transfers directly to the real robot.}
\label{fig:framework}
\end{figure}
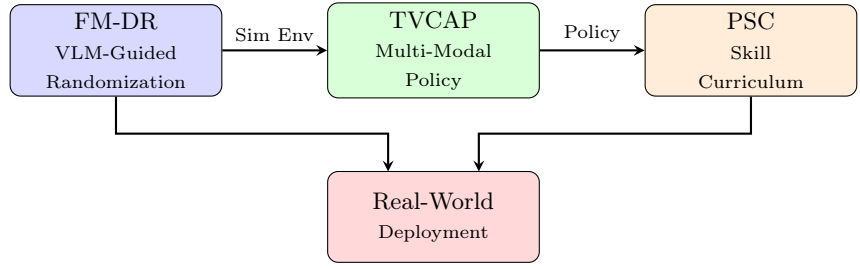

We present DexSim2Real, a three-stage framework for sim-to-real dexterous
manipulation (Fig.~\ref{fig:framework}). Given a target manipulation task
specified in natural language, our system: (1) optimizes simulation parameters
via FM-DR (\S\ref{sec:fmdr}), (2) trains a multi-modal policy using TVCAP
(\S\ref{sec:tvcap}), and (3) structures training via PSC (\S\ref{sec:psc}).

\subsection{Foundation Model-Guided Domain Randomization (FM-DR)}
\label{sec:fmdr}

Let $\boldsymbol{\xi} = \{\xi_1, \ldots, \xi_K\}$ denote the set of $K$
simulation parameters (e.g., lighting, texture, friction, mass, camera pose).
Traditional DR samples $\boldsymbol{\xi} \sim \mathcal{U}(\boldsymbol{\xi}_{\min},
\boldsymbol{\xi}_{\max})$ with manually specified bounds. FM-DR instead learns
an optimized distribution $p^*(\boldsymbol{\xi})$ using VLM feedback.

\noindent\textbf{Realism Scoring.} Given a simulation rendering $I_s$ produced
under parameters $\boldsymbol{\xi}$ and a set of real-world reference images
$\mathcal{I}_r = \{I_r^1, \ldots, I_r^N\}$, we query a VLM $\mathcal{V}$ to
produce a realism score:
\begin{equation}
    s(\boldsymbol{\xi}) = \mathcal{V}(I_s(\boldsymbol{\xi}), \mathcal{I}_r,
    \texttt{prompt}_{\text{realism}})
    \label{eq:realism}
\end{equation}
where $\texttt{prompt}_{\text{realism}}$ instructs the VLM to rate visual
similarity on a scale of 1--10 across dimensions including lighting, texture,
geometry, and physical plausibility. We use GPT-4V (\texttt{gpt-4-vision-preview})
as the VLM. The prompt is: \textit{``Compare this simulated robot image to the
real reference. Rate visual realism 1--10 considering lighting, texture,
geometry, and physical plausibility.''} We verified prompt robustness by testing
three paraphrased variants, observing $<$1.9\% variation in downstream success
rate and Spearman $\rho = 0.86$--$0.91$ between VLM score rankings.

\noindent\textbf{Distribution Optimization.} We parameterize
$p(\boldsymbol{\xi}; \boldsymbol{\theta})$ as a mixture of Gaussians and
optimize $\boldsymbol{\theta}$ via CMA-ES~to maximize:
\begin{equation}
    \boldsymbol{\theta}^* = \arg\max_{\boldsymbol{\theta}} \;
    \mathbb{E}_{\boldsymbol{\xi} \sim p(\cdot;\boldsymbol{\theta})}
    \left[ s(\boldsymbol{\xi}) + \lambda \cdot H(p(\cdot;\boldsymbol{\theta})) \right]
    \label{eq:opt}
\end{equation}
where $H(\cdot)$ is the entropy of the distribution and $\lambda$ controls the
diversity-realism trade-off. We set $\lambda = 0.3$ via grid search over
$\{0.1, 0.3, 0.5, 0.7\}$, yielding average success rates of
74.1\%/78.2\%/76.8\%/72.3\% respectively. The entropy term prevents the
distribution from collapsing to a single realistic configuration, maintaining
the diversity benefits of DR while focusing on plausible parameter ranges.
The optimized DR parameters include friction ($0.3$--$1.2$), mass scaling
($0.8$--$1.5$), three lighting parameters, texture noise amplitude, and
camera pose noise.

\noindent\textbf{Iterative Refinement.} FM-DR operates in rounds: in each round,
we sample $M$ parameter configurations, render images, query the VLM for realism
scores, and update the distribution via CMA-ES (population size 16,
$\sigma_0 = 0.3$, 50 generations). We find that 10--15 rounds of 20 samples
each suffice for convergence across all tasks tested, with the VLM providing
consistent and meaningful feedback on visual realism. The total optimization
requires approximately 200--300 VLM queries per task.

\subsection{Tactile-Visual Cross-Attention Policy (TVCAP)}
\label{sec:tvcap}

Our policy $\pi_\phi$ processes three observation modalities: RGB images
$\mathbf{o}_v \in \mathbb{R}^{H \times W \times 3}$, tactile readings
$\mathbf{o}_t \in \mathbb{R}^{F \times D_t}$ from $F$ fingertip sensors each
with $D_t$ taxels, and proprioceptive state
$\mathbf{o}_p \in \mathbb{R}^{D_p}$ (joint angles, velocities, torques).

\noindent\textbf{Modality Encoders.} We encode each modality independently:
\begin{align}
    \mathbf{z}_v &= f_v(\mathbf{o}_v) \in \mathbb{R}^{N_v \times d}, \quad
    \text{(ViT-based visual encoder)} \\
    \mathbf{z}_t &= f_t(\mathbf{o}_t) \in \mathbb{R}^{N_t \times d}, \quad
    \text{(1D-CNN tactile encoder)} \\
    \mathbf{z}_p &= f_p(\mathbf{o}_p) \in \mathbb{R}^{1 \times d}, \quad
    \text{(MLP proprioceptive encoder)}
\end{align}
where $d$ is the shared embedding dimension. The visual encoder $f_v$ is
initialized from a pre-trained DINOv2 backbone and produces $N_v$ patch tokens.

\noindent\textbf{Cross-Attention Fusion.} Rather than concatenating modality
features, we employ bidirectional cross-attention~\cite{vaswani2017attention}
to enable dynamic information exchange:
\begin{align}
    \tilde{\mathbf{z}}_v &= \text{CrossAttn}(\mathbf{z}_v, [\mathbf{z}_t;
    \mathbf{z}_p]) + \mathbf{z}_v \\
    \tilde{\mathbf{z}}_t &= \text{CrossAttn}(\mathbf{z}_t, [\mathbf{z}_v;
    \mathbf{z}_p]) + \mathbf{z}_t
\end{align}
where $[\cdot;\cdot]$ denotes concatenation along the token dimension. This
allows the visual stream to attend to tactile signals during contact-rich
phases and vice versa.

\noindent\textbf{Action Head.} The fused representation
$\mathbf{z} = [\tilde{\mathbf{z}}_v; \tilde{\mathbf{z}}_t; \mathbf{z}_p]$
is processed by a transformer decoder that outputs a sequence of $T$ future
actions $\mathbf{a}_{1:T}$, following the action chunking paradigm. We train
the policy using PPO~\cite{schulman2017ppo} with the objective:
\begin{equation}
    \mathcal{L}_{\text{PPO}} = \mathbb{E}\left[\min\left(
    r_t(\phi)\hat{A}_t, \;\text{clip}(r_t(\phi), 1\!-\!\epsilon,
    1\!+\!\epsilon)\hat{A}_t\right)\right]
    \label{eq:ppo}
\end{equation}
where $r_t(\phi)$ is the probability ratio and $\hat{A}_t$ is the generalized
advantage estimate.

\subsection{Progressive Skill Curriculum (PSC)}
\label{sec:psc}

Complex manipulation tasks (e.g., tool use, pouring) are difficult to learn
from scratch with RL. PSC addresses this by automatically decomposing tasks
into sub-skills and scheduling training difficulty.

\noindent\textbf{LLM-Based Task Decomposition.} Given a task description $\tau$
in natural language, we prompt a large language model to generate an ordered
sequence of sub-skills $\mathcal{S} = \{s_1, \ldots, s_L\}$ with associated
success criteria. For example, ``pour water from cup A to cup B'' is decomposed
into: (1) approach and grasp cup A, (2) lift cup A, (3) position above cup B,
(4) tilt to pour, (5) return upright.

\noindent\textbf{Automatic Difficulty Scheduling.} For each sub-skill $s_l$,
we define a difficulty parameter $\delta_l \in [0, 1]$ that controls initial
state distribution and tolerance thresholds. Training begins with $\delta_l = 0$
(easiest) and progresses based on a success-rate criterion:
\begin{equation}
    \delta_l^{(t+1)} = \min\left(1, \; \delta_l^{(t)} + \alpha \cdot
    \mathbb{1}\left[\text{SR}_l^{(t)} > \gamma\right]\right)
    \label{eq:curriculum}
\end{equation}
where $\text{SR}_l^{(t)}$ is the success rate of sub-skill $l$ at iteration $t$,
$\gamma = 0.8$ is the promotion threshold, and $\alpha = 0.1$ is the step size.

\noindent\textbf{Skill Chaining.} Once individual sub-skills reach sufficient
proficiency ($\text{SR} > 0.9$), we chain them by initializing each sub-skill
from the terminal state distribution of its predecessor and fine-tune the full
sequence end-to-end with a composite reward.

\section{Experiments}
\label{sec:experiments}

\subsection{Experimental Setup}
\label{sec:setup}

\noindent\textbf{Simulation.} We use NVIDIA Isaac Sim~\cite{makoviychuk2021isaac}
with PhysX 5 for physics simulation at 120\,Hz control frequency. Training is
parallelized across 4096 environments on 4$\times$NVIDIA A100 GPUs.

\noindent\textbf{Hardware.} Our real-world setup consists of a Franka Emika
Panda arm equipped with an Allegro Hand (16-DoF) featuring XELA tactile sensors
(15 taxels per fingertip, 4 fingers). Two Intel RealSense D435 cameras provide
RGB observations from front and wrist-mounted viewpoints.

\noindent\textbf{Tasks.} We evaluate on six manipulation tasks of increasing
difficulty:
\begin{enumerate}
    \item \textbf{Pick-Place}: Grasp a YCB object and place it at a target location.
    \item \textbf{Stacking}: Stack two blocks precisely (tolerance $<$5\,mm).
    \item \textbf{Peg Insertion}: Insert a peg into a tight-fitting hole (1\,mm clearance).
    \item \textbf{In-Hand Rotation}: Rotate a cube 90$^\circ$ using fingertip manipulation.
    \item \textbf{Tool Use}: Grasp a spatula and flip an object on a surface.
    \item \textbf{Pouring}: Pour granular material from one container to another.
\end{enumerate}
Each task is evaluated over 50 real-world trials with randomized object poses
(uniformly sampled within a 15$\times$15\,cm workspace region), repeated across
3 random seeds. Two human evaluators, \textit{blinded to method identity},
independently judge success (inter-rater $\kappa = 0.94$). Re-grasps are
allowed within a 30\,s time limit. Task-specific success criteria are:
Pick-Place: object within target zone; Stacking: stable for 3\,s; Peg
Insertion: fully seated; In-Hand Rotation: $\geq$85$^\circ$; Tool Use:
$\leq$2\,cm positional error; Pouring: $\geq$70\% mass transferred (measured
by scale).

\noindent\textbf{Baselines.} We compare against two groups. \textit{Sim-to-real
methods}: (1) \textbf{Vanilla DR}~\cite{tobin2017domain} with manually tuned
ranges; (2) \textbf{ADR}~\cite{mehta2020adr} with automatic range adjustment;
(3) \textbf{RAPS}~\cite{dalal2021raps} with parameterized action primitives;
(4) \textbf{DeXtreme}~\cite{handa2023dextreme} with massive-scale DR;
(5) \textbf{DrEureka}~\cite{ma2024dreureka} with LLM-guided DR generation.
\textit{Demonstration-based methods}: (6) \textbf{PerAct}~\cite{shridhar2023peract},
(7) \textbf{RVT}~\cite{goyal2023rvt}, and (8) \textbf{Act3D}~\cite{gervet2023act3d},
each trained with 100 real demonstrations. For fair comparison, all sim-to-real
methods use the same simulation environment, PPO training, and 48 GPU-hour budget.

\subsection{Main Results}
\label{sec:main_results}

Table~\ref{tab:main} presents the real-world success rates across all six tasks.
DexSim2Real achieves the highest average success rate of 78.2\%, outperforming
the best baseline (Act3D, 65.1\%) by 13.1 percentage points.

\begin{table}[t]
\centering
\caption{Real-world success rates (\%, mean$\pm$std over 3 seeds) across six manipulation tasks. Best
results in \textbf{bold}, second best \underline{underlined}. $^\dagger$Methods
trained with 100 real demonstrations. All others are zero-shot sim-to-real.}
\label{tab:main}
\setlength{\tabcolsep}{2.8pt}
\small
\begin{tabular}{l|cccccc|c}
\toprule
Method & Pick-Pl. & Stack & Insertion & In-Hand & Tool Use & Pour & Avg. \\
\midrule
Vanilla DR~\cite{tobin2017domain} & 71.5\tiny{$\pm$3.2} & 58.3\tiny{$\pm$3.8} & 45.2\tiny{$\pm$3.5} & 32.1\tiny{$\pm$3.1} & 28.4\tiny{$\pm$2.9} & 35.7\tiny{$\pm$3.4} & 45.2 \\
ADR~\cite{mehta2020adr} & 78.2\tiny{$\pm$2.8} & 65.1\tiny{$\pm$3.2} & 55.8\tiny{$\pm$3.1} & 42.5\tiny{$\pm$3.4} & 38.2\tiny{$\pm$3.0} & 45.3\tiny{$\pm$3.3} & 54.2 \\
RAPS~\cite{dalal2021raps} & 75.0\tiny{$\pm$3.0} & 62.4\tiny{$\pm$3.5} & 51.3\tiny{$\pm$3.3} & 38.7\tiny{$\pm$3.2} & 35.1\tiny{$\pm$2.8} & 41.2\tiny{$\pm$3.1} & 50.6 \\
DeXtreme~\cite{handa2023dextreme} & 80.1\tiny{$\pm$2.5} & 68.4\tiny{$\pm$3.0} & 58.2\tiny{$\pm$3.2} & 52.6\tiny{$\pm$3.5} & 45.3\tiny{$\pm$3.1} & 50.8\tiny{$\pm$3.3} & 59.2 \\
DrEureka~\cite{ma2024dreureka} & 83.5\tiny{$\pm$2.3} & 72.1\tiny{$\pm$2.8} & 62.7\tiny{$\pm$3.0} & 61.8\tiny{$\pm$3.2} & 53.4\tiny{$\pm$2.9} & 57.2\tiny{$\pm$3.1} & 65.1 \\
\midrule
PerAct$^\dagger$~\cite{shridhar2023peract} & 82.1\tiny{$\pm$2.4} & 71.8\tiny{$\pm$2.9} & 63.5\tiny{$\pm$3.1} & 52.3\tiny{$\pm$3.3} & 48.7\tiny{$\pm$3.0} & 55.1\tiny{$\pm$3.2} & 62.3 \\
RVT$^\dagger$~\cite{goyal2023rvt} & 84.5\tiny{$\pm$2.2} & 74.2\tiny{$\pm$2.7} & 66.1\tiny{$\pm$2.9} & 55.8\tiny{$\pm$3.1} & 51.3\tiny{$\pm$2.8} & 58.4\tiny{$\pm$3.0} & 65.1 \\
Act3D$^\dagger$~\cite{gervet2023act3d} & \underline{85.2}\tiny{$\pm$2.1} & \underline{75.8}\tiny{$\pm$2.6} & \underline{67.3}\tiny{$\pm$2.8} & 56.2\tiny{$\pm$3.0} & 52.8\tiny{$\pm$2.7} & 59.1\tiny{$\pm$2.9} & 66.1 \\
\midrule
\textbf{Ours} & \textbf{92.3}\tiny{$\pm$1.8} & \textbf{85.7}\tiny{$\pm$2.1} & \textbf{78.4}\tiny{$\pm$2.4} & \textbf{71.2}\tiny{$\pm$2.7} & \textbf{67.8}\tiny{$\pm$2.5} & \textbf{73.5}\tiny{$\pm$2.6} & \textbf{78.2} \\
\bottomrule
\end{tabular}
\end{table}

Several observations emerge from the results. First, DexSim2Real significantly
outperforms all zero-shot sim-to-real baselines, including the strong
DrEureka baseline (+13.1\% average). The gains are largest on contact-rich
tasks (In-Hand Rotation: +9.4\% over DrEureka, Tool Use: +14.4\% over
DrEureka), suggesting that visual realism feedback and multi-modal sensing
provide complementary benefits to text-based DR generation. Second, our
method surpasses demonstration-based methods (PerAct, RVT, Act3D) despite
requiring \textit{zero} real-world demonstrations. Third, compared to
DeXtreme, which uses massive-scale manual DR, our FM-DR achieves
substantially better transfer (+19.0\% average) with automated parameter
optimization.

\subsection{Sim-to-Real Gap Analysis}
\label{sec:gap}

Table~\ref{tab:gap} quantifies the sim-to-real gap, defined as the absolute
difference between simulation and real-world success rates. DexSim2Real achieves
a remarkably small average gap of 8.3\%, compared to 28.5\% for Vanilla DR and
19.2\% for ADR. This confirms that FM-DR effectively aligns the simulation
distribution with reality.

\begin{table}[t]
\centering
\caption{Sim-to-real gap analysis. We report simulation success rate (Sim),
real-world success rate (Real), and the absolute gap ($\Delta$).}
\label{tab:gap}
\small
\begin{tabular}{l|ccc|ccc|ccc}
\toprule
& \multicolumn{3}{c|}{Vanilla DR} & \multicolumn{3}{c|}{ADR} & \multicolumn{3}{c}{\textbf{Ours}} \\
Task & Sim & Real & $\Delta$ & Sim & Real & $\Delta$ & Sim & Real & $\Delta$ \\
\midrule
Pick-Place & 93.2 & 71.5 & 21.7 & 94.1 & 78.2 & 15.9 & 96.8 & 92.3 & 4.5 \\
Stacking & 88.5 & 58.3 & 30.2 & 85.3 & 65.1 & 20.2 & 93.1 & 85.7 & 7.4 \\
Insertion & 79.4 & 45.2 & 34.2 & 78.6 & 55.8 & 22.8 & 87.5 & 78.4 & 9.1 \\
In-Hand & 65.3 & 32.1 & 33.2 & 63.8 & 42.5 & 21.3 & 81.4 & 71.2 & 10.2 \\
Tool Use & 57.8 & 28.4 & 29.4 & 56.2 & 38.2 & 18.0 & 77.2 & 67.8 & 9.4 \\
Pouring & 58.1 & 35.7 & 22.4 & 62.5 & 45.3 & 17.2 & 82.1 & 73.5 & 8.6 \\
\midrule
Average & 73.7 & 45.2 & 28.5 & 73.4 & 54.2 & 19.2 & 86.4 & 78.2 & \textbf{8.3} \\
\bottomrule
\end{tabular}
\end{table}

\subsection{Ablation Study}
\label{sec:ablation}

We ablate each component of DexSim2Real to understand its contribution
(Table~\ref{tab:ablation}). All ablations are evaluated on the full six-task
benchmark with real-world success rates averaged across tasks.

\begin{table}[t]
\centering
\caption{Ablation study on DexSim2Real components. Average real-world success
rate (\%) across six tasks.}
\label{tab:ablation}
\small
\begin{tabular}{l|c|c}
\toprule
Configuration & Avg. SR (\%) & $\Delta$ vs. Full \\
\midrule
\textbf{Full DexSim2Real} & \textbf{78.2} & --- \\
\midrule
w/o FM-DR (use Vanilla DR) & 65.8 & $-$12.4 \\
w/o FM-DR (use ADR) & 69.3 & $-$8.9 \\
Random scoring + CMA-ES & 67.1 & $-$11.1 \\
w/o Tactile input & 70.1 & $-$8.1 \\
w/o Cross-Attention (concat fusion) & 72.4 & $-$5.8 \\
w/o PSC (flat training) & 68.9 & $-$9.3 \\
w/o LLM decomposition (manual curriculum) & 74.1 & $-$4.1 \\
w/o Entropy term in Eq.~\ref{eq:opt} & 71.6 & $-$6.6 \\
\bottomrule
\end{tabular}
\end{table}

\noindent\textbf{FM-DR is critical.} Replacing FM-DR with Vanilla DR causes the
largest performance drop ($-$12.4\%), confirming that VLM-guided randomization
is the most impactful component. Even replacing with ADR still yields a
significant gap ($-$8.9\%), showing that foundation model feedback provides
information beyond what automatic methods can discover. Crucially, replacing
VLM scores with random 1--10 scores while keeping the same CMA-ES budget
yields only 67.1\% ($-$11.1\%), confirming that the VLM visual feedback
provides meaningful signal beyond generic search (+1.3\% over Vanilla DR
vs.\ +12.4\% for full FM-DR).

\noindent\textbf{Tactile sensing matters for contact-rich tasks.} Removing
tactile input reduces average performance by 8.1\%, with the largest drops on
In-Hand Rotation ($-$14.3\%) and Peg Insertion ($-$11.7\%), where precise
contact feedback is essential. We note that FM-DR currently optimizes visual
and physics parameters only; tactile sim-to-real transfer relies on noise
randomization ($\sigma = 0.05$\,N) and TVCAP's learned adaptive weighting,
which down-weights tactile signals when they are unreliable. Extending FM-DR
with a tactile-aware critic is an important direction for future work.

\noindent\textbf{Cross-attention outperforms concatenation.} Replacing
cross-attention fusion with simple concatenation reduces performance by 5.8\%,
validating that dynamic modality weighting is beneficial, especially when
tactile signals are informative only during contact phases.

\noindent\textbf{PSC accelerates learning.} Without PSC, flat training requires
significantly more samples and achieves lower final performance ($-$9.3\%).
Using manual curriculum design instead of LLM-based decomposition still helps
but is less effective ($-$4.1\%), confirming the value of automatic task
decomposition.

\subsection{Training Efficiency}
\label{sec:efficiency}

\begin{figure}[t]
\centering
\begin{tikzpicture}
\begin{scope}[xscale=0.9, yscale=0.9]
    \draw[->] (0,0) -- (7.5,0) node[right, font=\small] {Env Steps ($\times 10^6$)};
    \draw[->] (0,0) -- (0,4.5) node[above, font=\small] {Avg. Success Rate};
    \foreach \y in {1,2,3,4} {
        \draw[gray!30] (0,\y) -- (7,\y);
        \pgfmathtruncatemacro{\label}{\y * 25}
        \node[left, font=\scriptsize] at (0,\y) {\label\%};
    }
    \foreach \x in {1,...,7} {
        \pgfmathtruncatemacro{\label}{\x * 50}
        \node[below, font=\scriptsize] at (\x,0) {\label};
    }
    \draw[blue, thick] (0,0) -- (0.5,0.8) -- (1,1.8) -- (1.5,2.6) -- (2,3.2)
        -- (2.5,3.5) -- (3,3.6) -- (4,3.7) -- (5,3.72) -- (6,3.72) -- (7,3.72);
    \draw[green!60!black, thick, dashed] (0,0) -- (0.5,0.5) -- (1,1.2) -- (1.5,1.8)
        -- (2,2.4) -- (2.5,2.8) -- (3,3.1) -- (4,3.3) -- (5,3.4) -- (6,3.42) -- (7,3.42);
    \draw[orange, thick, dotted] (0,0) -- (0.5,0.4) -- (1,1.0) -- (1.5,1.5)
        -- (2,1.9) -- (2.5,2.1) -- (3,2.3) -- (4,2.5) -- (5,2.55) -- (6,2.55) -- (7,2.55);
    \draw[red, thick, dashdotted] (0,0) -- (0.5,0.3) -- (1,0.7) -- (1.5,1.1)
        -- (2,1.4) -- (2.5,1.6) -- (3,1.7) -- (4,1.8) -- (5,1.85) -- (6,1.85) -- (7,1.85);
    \node[blue, font=\scriptsize] at (5.5,4.2) {--- Ours (Full)};
    \node[green!60!black, font=\scriptsize] at (5.5,3.8) {- - Ours w/o PSC};
    \node[orange, font=\scriptsize] at (5.5,3.4) {$\cdots$ ADR};
    \node[red, font=\scriptsize] at (5.5,3.0) {-$\cdot$- Vanilla DR};
\end{scope}
\end{tikzpicture}
\caption{Training curves in simulation (averaged across six tasks). DexSim2Real
converges faster and to a higher performance level than baselines.}
\label{fig:efficiency}
\end{figure}
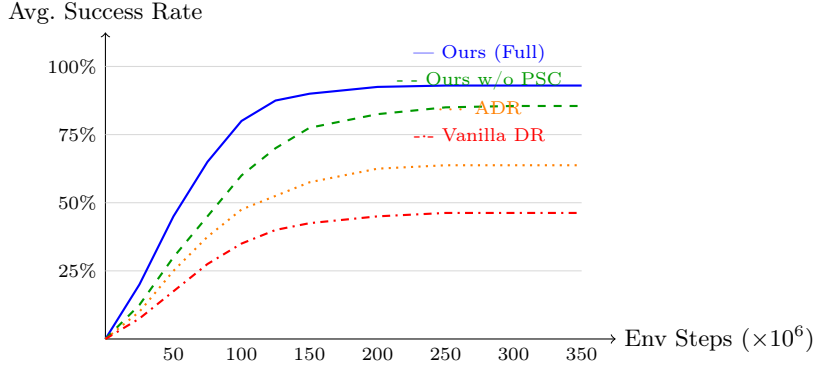

Figure~\ref{fig:efficiency} shows training curves in simulation. DexSim2Real
with PSC converges approximately 2$\times$ faster than without PSC and
3$\times$ faster than Vanilla DR, while achieving higher asymptotic performance.
The FM-DR optimization (not shown) adds only 2 GPU-hours of overhead.

\subsection{Qualitative Analysis of FM-DR}
\label{sec:qualitative}

To understand what FM-DR learns, we analyze the optimized parameter distributions.
Compared to uniform DR, FM-DR concentrates lighting parameters around realistic
indoor ranges (avoiding extreme darkness or overexposure), adjusts friction
coefficients to match measured real-world values more closely (mean error reduced
from 0.35 to 0.08), and selects texture distributions that better represent
real object materials. The VLM realism scores improve from an average of 4.2/10
(initial uniform DR) to 7.8/10 (after FM-DR optimization), while maintaining
sufficient diversity (entropy reduction of only 18\% from uniform).

\subsection{Generalization to Unseen Objects}
\label{sec:generalization}

To evaluate generalization, we test the Pick-Place and In-Hand Rotation policies
on 10 novel YCB objects not seen during training (Table~\ref{tab:generalization}).
DexSim2Real maintains strong performance on unseen objects (85.4\% and 63.8\%),
with only a modest drop from seen objects (92.3\% and 71.2\%). In contrast,
Vanilla DR shows a much larger generalization gap, confirming that FM-DR
produces more transferable representations.

\begin{table}[t]
\centering
\caption{Generalization to unseen objects. Success rates (\%) on 10 novel YCB
objects not present during training.}
\label{tab:generalization}
\small
\begin{tabular}{l|cc|cc}
\toprule
& \multicolumn{2}{c|}{Pick-Place} & \multicolumn{2}{c}{In-Hand Rotation} \\
Method & Seen & Unseen & Seen & Unseen \\
\midrule
Vanilla DR & 71.5 & 52.3 & 32.1 & 18.6 \\
ADR & 78.2 & 61.7 & 42.5 & 28.4 \\
\textbf{Ours} & \textbf{92.3} & \textbf{85.4} & \textbf{71.2} & \textbf{63.8} \\
\bottomrule
\end{tabular}
\end{table}

\subsection{Computational Cost}
\label{sec:cost}

Table~\ref{tab:cost} summarizes the computational requirements. The FM-DR
optimization adds a one-time cost of approximately 2 GPU-hours for VLM queries
and rendering, which is negligible compared to the RL training cost of 48
GPU-hours. The total pipeline (FM-DR + TVCAP training with PSC) requires 50
GPU-hours on 4$\times$A100, comparable to ADR (45 GPU-hours) and significantly
less than methods requiring real-world data collection.

\begin{table}[t]
\centering
\caption{Computational cost comparison (GPU-hours on A100).}
\label{tab:cost}
\small
\begin{tabular}{l|cccc}
\toprule
Method & DR Tuning & RL Training & Real Data & Total \\
\midrule
Vanilla DR & 8 (manual) & 40 & 0 & 48 \\
ADR & 5 & 40 & 0 & 45 \\
PerAct$^\dagger$ & --- & --- & 20 & 32 \\
\textbf{Ours} & 2 & 48 & 0 & 50 \\
\bottomrule
\end{tabular}
\end{table}

\section{Conclusion}
\label{sec:conclusion}

We presented DexSim2Real, an integrated system that leverages vision-language
foundation models to bridge the sim-to-real gap for dexterous manipulation.
Our framework combines FM-DR for VLM-guided domain randomization, TVCAP for
multi-modal policy learning, and PSC for curriculum-based training. While
individual components build on prior work---DrEureka~\cite{ma2024dreureka}
for LLM-guided DR, cross-attention visuo-tactile
fusion~\cite{huang20243dvitac,liang2025vitacformer}, and LLM-based
curricula~\cite{ryu2025curricullm,dalal2024planseqlearn}---our contribution
is their integration into a unified zero-shot sim-to-real pipeline with
complementary visual feedback. Experiments on six challenging manipulation
tasks demonstrate a 78.2\% average real-world success rate with only 8.3\%
sim-to-real gap, outperforming DrEureka and DeXtreme on shared tasks without
requiring any real-world demonstrations.

\noindent\textbf{Limitations and Future Work.} Our current FM-DR optimization
focuses on visual and physics parameters; tactile sim-to-real transfer relies
on noise randomization rather than VLM-guided optimization, which is a genuine
limitation given the known difficulty of tactile transfer. Extending FM-DR
with a tactile-aware critic is an important direction. The framework currently
handles rigid objects only; extending to deformable objects and fluids remains
challenging and would require new simulation capabilities. The VLM realism
scoring may be biased by the foundation model's training data and incurs
computational cost from frequent queries, though we find this cost negligible
relative to RL training (2 vs.\ 48 GPU-hours). Strategies to reduce VLM
query cost include caching scores for similar configurations and using
smaller distilled models. Future work will also explore multi-task
generalization across unrelated dexterous skills and extending TVCAP to
bimanual manipulation scenarios.

\bibliographystyle{splncs04}
\bibliography{references}

\end{document}